\documentclass[conf]{new-aiaa}
\usepackage[utf8]{inputenc}

\usepackage{graphicx}
\usepackage{amsmath}
\usepackage{textcomp}
\usepackage[version=4]{mhchem}
\usepackage{siunitx}
\usepackage{wrapfig}
\usepackage{longtable,tabularx}
\usepackage{booktabs,caption}
\usepackage[flushleft]{threeparttable}
\usepackage[flushleft]{threeparttable}
\usepackage{url}

\title{Training Environment for High Performance \\Aircraft Reinforcement Learning}

\author{Gregory F. Search\footnote{Experimental RPA Test Pilot, 452 FLTS}}

\begin{document}

\maketitle

\begin{abstract}
This paper presents Tunnel, a simple, open source, reinforcement learning training environment for high performance aircraft. It integrates the F-16's 3D non-linear flight dynamics into OpenAI's Gymnasium python package. The template includes primitives for boundaries, targets, adversaries and sensing capabilities that may vary depending on operational need. This offers mission planners a means to rapidly respond to evolving environments, sensor capabilities and adversaries for autonomous air combat aircraft. It offers researchers access to operationally relevant aircraft physics. Tunnel's simple code base is accessible to anyone familiar with Gymnasium and/or those with basic python skills. This paper includes  a demonstration of a week long trade study that investigated a variety of training methods, observation spaces, and threat presentations. This enables increased collaboration between researchers and mission planners which can translate to a national military advantage. As warfare becomes increasingly reliant upon automation, software agility will correlate with decision advantages. Airmen must have tools to adapt to adversaries in this context. It may take months for researchers to develop skills to customize observation, actions, tasks and training methodologies in air combat simulators. In Tunnel, this can be done in a matter of days. 

\end{abstract}

\section{Nomenclature}

{\renewcommand\arraystretch{1.0}
\noindent\begin{longtable*}{@{}l @{\quad=\quad} l@{}}
$AESA$ & Active Electronically Scanned Array\\
$BFM$ & Basic Fighter Maneuvers \\
$CCA$ & Collaborative Combat Aircraft \\
$CNN$  & Convolutional Neural Network\\
$CSAIL$ & Computer Science and Artificial Intelligence Lab \\
$DARPA$ & Defense Advanced Research Projects Agency \\
$EW$ & Electronic Warfare \\
$F-16$ & Designation for the Fighting Falcon Aircraft \\
$FLCS$ & Flight Control System \\
$GPS$ & Global Positioning System \\
$LiDAR$ & Light Detection and Ranging \\
$LSTM$ & Long Short Term Memory \\
$LTC$ & Liquid Time Constant \\
$MIT$ & Massachusetts Institute of Technology \\
$ML$ & Machine Learning \\
$MLP$ & Multi-Layer Perceptron \\  
$PID$ & Proportional Integrator Differentiator \\
$RNN$ & Recurrent Neural Network \\
$RWR$ & Radar Warning Receiver \\
$UAV$ & Unmanned Aerial Vehicle \\
$VENOM$ & Viper Experimental Next-Gen Operations Model \\
$VISTA$ & Variable In-Flight Stability Test Aircraft

\end{longtable*}}

\section{Introduction}
\lettrine{F}{or} the first time in over sixty years, the United States Air Force's capability to achieve air superiority is at risk \cite{mitchellinstitute}. To mitigate this, Collaborative Combat Aircraft (CCA) are being designed with levels of autonomy never before seen in high performance air combat \cite{CCA}. The state of the art is an F-16 capable of being controlled by an AI agent named the Variable In-Flight Stability Test Aircraft (VISTA). Though it can perform live air-air engagements, this unique aircraft design requires funds and time that are prohibitive to plans to deliver over one thousand CCA. Furthermore, VISTA does not integrate real world sensors, contested environments, non-cooperative adversaries or missions beyond air-air. The nation needs a way to rapidly discover capabilities and limitations of autonomous agents in a variety of air combat situations. This paper introduces Tunnel, a reinforcement learning environment created by the researcher. It is designed for simple modification to present the agent with a range of observations, actions, tasks and training methodologies. 
\par 
Currently, assessing autonomy in the high performance air domain is a slow process. The most recent VISTA flights have been in support of DARPA Air Combat Evolution (ACE)\cite{ACE}. Teams undergo a rigorous build up from software in the loop simulation, then hardware in the loop simulation and constructive modeling prior to live flight. In 2022, the Department of the Air Force - Massachusetts Institute of Technology Artificial Intelligence Accelerator (DAF-MIT AI Accelerator, or DAF AIA) formed a team to participate in DARPA ACE. This team was able to train a novel class of algorithm, called a Liquid Time-constant Network \cite{LTC}, to perform live autonomous flight within six months. This was less time than ACE training timelines, which can take years. Often times, teams must sacrifice exploration of algorithms to meet programmatic constraints imposed by this training timeline. Furthermore, the build-up to flight presented the teams with a fixed observation, action, task and training methodology. As autonomy evolves to handle real sensors, diverse mission environments and non-cooperative adversaries, these programmatic constraints may worsen if not properly mitigated. \par
In the coming years, air combat autonomy will need to operate in far more challenging environments. The lessons learned from DARPA ACE are expected to be used to advance DARPA's AI Reinforcements (AIR) project as well as the US Air Force's Viper Experimental Next-Gen Operations Model (VENOM) program. AIR plans to research autonomous air combat with multiple agents while subjected to partial observability, concept drift and uncertainty \cite{AIR}. VENOM plans to use operationally configured F-16 aircraft as high performance airborne test beds \cite{VENOM}. The intent is to use data ingested from operational sensors to construct an observation for the agent.

\section{Related Work}
There are already simulations that can present AI agents with different air combat situations. There are also training environments that allow for rapid, diverse exploration of capabilities and limitations. Tunnel seeks to achieve the flexibility of training environments while maintaining a relevant representation of the high performance air domain. Some of these efforts are listed below:\\ \\
\textbf{Drone Simulations} - Airsim \cite{airsim} and Pybullet-drones \cite{pybullet} are examples of training environments that model the behavior of small UAV. Some are effective at simulating multi-drone configurations. However, many don't allow customization of the state, action and reward of the agent. Even if state, action and reward were to match, speed and maneuverability would likely make conclusions about high performance aircraft based on drone simulations invalid.\\
\textbf{DCS World and FSX} - Many software products offer access to the flight dynamics of the F-16 and other high performance aircraft. Digital Combat Simulator (DCS) world \cite{DCS} and Flight Simulator-X \cite{FSX} are both low cost and offer mission customization. While users can make modifications, implementation of AI would typically be effects based. Meaning, the developer cannot alter the physics of the game directly.\\ \textbf{Proprietary Simulations} - Defense primes such as Lockheed Martin, Northrop Grumman or Boeing own powerful, high fidelity simulations. These are prohibitive to most research teams due to distribution restrictions. Even for those allowed to use these products, there can be significant time and expertise required to learn to use the tool, let alone change the simulation environment.\\ 
\textbf{JSBSim and HarFang3D} - JSBSim is an open source flight dynamics model. It leverages OpenAI Gymnasium and includes dozens of aircraft models \cite{JSBSim}. Another example of open source simulation for high performance aircraft is HarFang3D, which features customizable air-to-air scenarios for reinforcement learning (RL) \cite{harfang}. Both of these are written in C++. While these efforts have a strong body of documentation and tutorial videos, Tunnel should not require detailed instruction to use and is written using three files and less than 300 lines of code. \\
\textbf{MuJoCo \& ALE} - Open AI's suite of RL environments have a strong diversity of tasks. MuJoCo challenges the agent to learn high dimensional observation and action spaces as a means to control the agents' movements \cite{MuJoCo}. The Arcade Learning Environment (ALE) has hundreds of games designed to explore reward optimization in a breadth of contexts \cite{ALE}. However, neither integrate the control laws of high performance aircraft. \\
\textbf{CoinRun} - This effort exposes the agent to a range of procedurally generated "levels" to quantify and encourage generalization \cite{CoinRun}. These are also open source. None of these iterations are tailored to the air domain.\\
In summary, these efforts are typically scoped to either the researcher or the operator. Tunnel endeavors to bridge the two. Its simple implementation of representative high performance dynamics and high stakes tasks enables close integration between these two stakeholders. 

\section{Background}
\subsection{Air Combat Autonomy}
Theories for autonomous air combat have existed since the early 1950s \cite{autoaircombat1950}. As computing resources have become more powerful, many efforts to solve air combat analytically have been conducted at increasing levels of fidelity \cite{autoaircombat1988} \cite{autoaircombat2010}. For the purpose of this paper, we define a high performance aircraft as one that has sufficient speed and maneuverability to legitimately participate in a world class Air-to-Air engagement. At present, only the AI enabled F-16 "VISTA" has demonstrated live high performance autonomous air combat \cite{vista}. \par
These flights require a rigorous build-up from software to hardware integration prior to mission execution. In the case of DARPA ACE, this involves coordination with dozens of organizations and months of detailed planning. Each team must learn the software associated with the project to begin training and evaluating models, which often takes months. In addition to the millions of dollars of flight time, this project requires resources from USAF Test Pilot School at Edwards AFB. If the CCA program is to be successful, air combat autonomy must evolve from a single, hand crafted F-16 aircraft to hundreds of affordable aircraft. \par
\begin{figure}[h!]
  \includegraphics[scale=0.5]{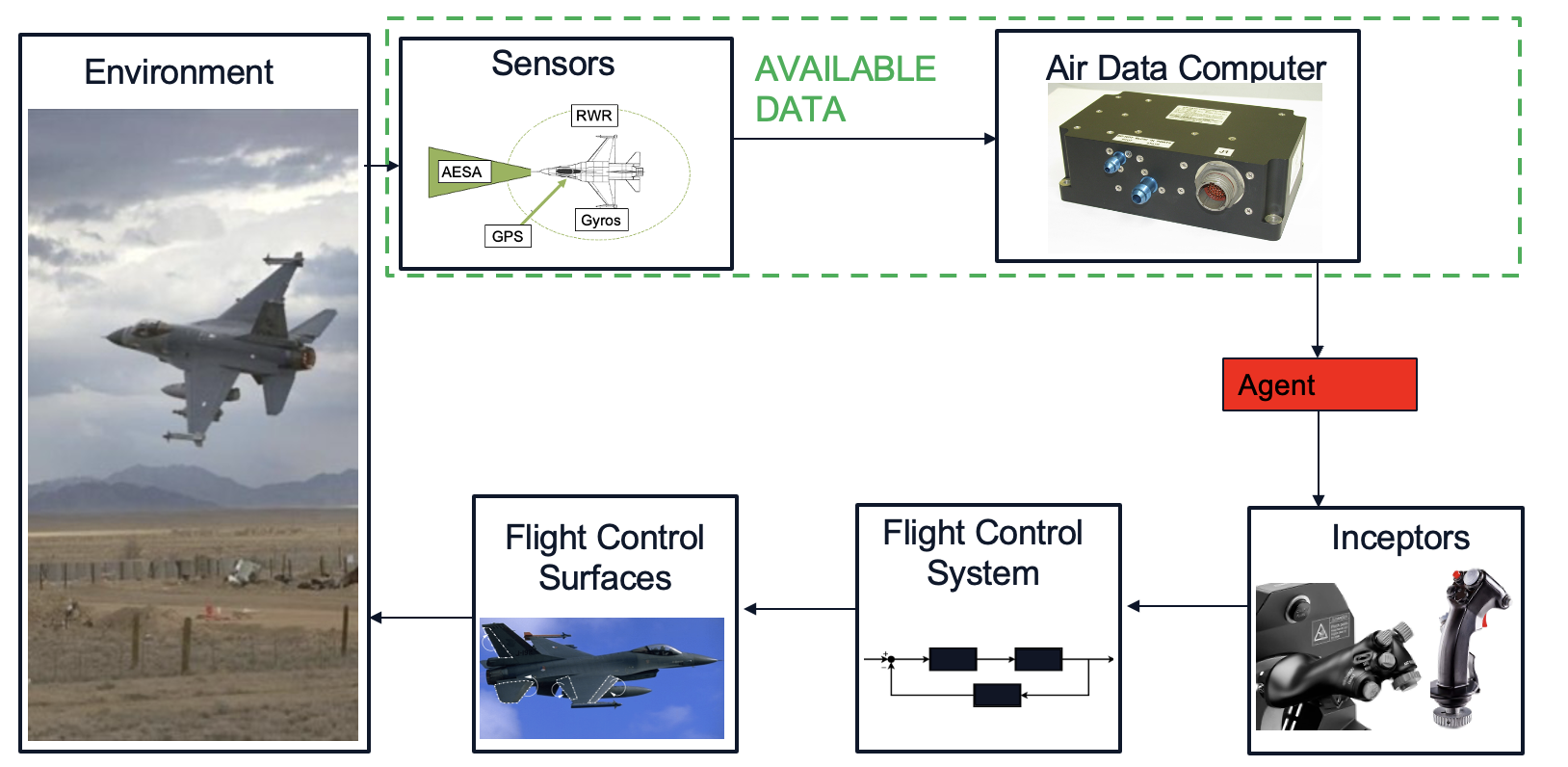}
  \caption{Human and Agent Executed Air Combat}
  \label{fig:whereami}
\end{figure}

In addition to the logistics capability gap between VISTA and CCA, there is a difference between VISTA environment and the environment a CCA will likely operate in real air operations \cite{vistadetailed}. Figure \ref{fig:whereami} provides a general pathway in which information is transferred from the environment to the operator, whose action leads to a change in the aircraft state within that environment. The air data computer can provide information to the agent processed at various levels.

Note the differences between the VISTA presentation and what would be presented operationally. Instead of receiving information from sensors, the agent receives all requisite data from the adversary. This sensor data could include megabits per second for the Air Data Computer to process. By comparison, the adversary state may include less than 100 parameters. For this reason, it will be non-trivial for designers of future autonomy platforms to decide how sensor information will be processed prior to presentation to the agent as an observation. This may include a variety of signal processing techniques. The action space has a similar level of design freedom. It has been shown that direct control of the flight control surfaces by an agent is not likely to succeed \cite{flcsdarkside}. However, the agent could be expected to perform high level actions such target prioritization. Agents could also be given control of stick, rudder and throttle. The latter was the case for DARPA ACE, but future mission may demand a different action space. In short, observation and action spaces in live air combat will be far more diverse than those presented in DARPA ACE. 

\subsection{F-16 Dynamics}
\begin{wrapfigure}[13]{l}{0.5\textwidth}
    \centering
    \includegraphics[width=0.5\textwidth]{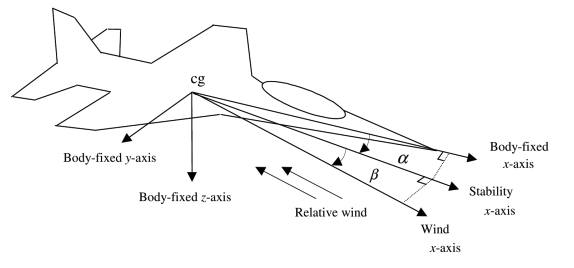}
    \caption{Axes Conventions, credit Heidlauf et al.~\cite{verification}}
    \label{fig:axes}
\end{wrapfigure}

Achieving sufficient control of high performance aircraft often involves increasing the complexity of the flight control system. This section gives an overview of the F-16 dynamics, which were used in DARPA ACE as well as in the Tunnel environment. To improve maneuverability, the F-16 airframe was designed to have relaxed static stability \cite{flcsdarkside}. This means that without augmentation, the aircraft will not return to a trimmed state after deviation. This Flight Control System (FLCS) provides this support to make the aircraft flyable by a process called "fly-by-wire" \cite{flybywire}. It has been said that the pilot doesn't actually fly the F-16 directly but instead provides a "request" to the FLCS. A useful model for F-16 dynamics is a 6 degree of freedom (DOF) model which provides the aircraft response to forces and moments in the x, y and z direction. As can be seen in Figure \ref{fig:axes}, these axes correspond to the perpendicular lines about which the aircraft rolls, pitches and yaws respectively.

These 6-DOF models can be constructed in a variety of ways by making assumptions which may include: flat Earth, inertial reference plane, rigid body, and more \cite{brandteom}. The model used in this paper uses an equation to describe forces, moments, kinematics and position of the aircraft in x, y, and z plus a thrust lag for a total of 13 equations \cite{verification}. Note that the x, y and z axis are "body-fixed," meaning as the aircraft rotates and translates, these axes remain constant with respect to the aircraft orientation. \par
The convention used in this paper and the code provided for Tunnel is "elevation" and "azimuth." Not to be confused with altitude, elevation refers to body-fixed angular increase from x axis. In other words, it is the positive degrees up from the perspective of the pilot. Azimuth refers to body-fixed left and right (+) degrees from the perspective of the pilot. 

\subsection{Algorithms and Training Methods} 
There are a range of approaches to automate aircraft control. In 1912, the Sperry Corporation developed the first aircraft autopilot which used heading and attitude measurements to operate the hydraulic elevator and rudder \cite{firstautopilot}. Modern autopilots for large aircraft can use signal processing to correct deviations from desired roll, pitch, yaw, latitude, longitude and altitude. However for high-speed, highly maneuverable aircraft like the F-16, signal processing alone is not sufficient to handle the full range of tactically relevant maneuvers. Heidlauf et al. describe the compounding complexity of control systems required when designing a build up to an automatic ground collision avoidance system \cite{verification}. \par
Classical control techniques, such as a PID (Proportional – Integral – Derivative) controller, may be able to control the F-16 \cite{pidf16}. This method adds control inputs proportional to the error between desired aircraft state and actual. It includes integral and derivative of the error terms over time to fine tune the response. An example of this in the F-16 is the "Death Claw" automatic gun run \cite{deathclaw}. Although there was some effectiveness prosecuting air-ground targets, air-air targets was not a viable operational use case for Death Claw. \par
Imitation learning has many encouraging possibilities in this domain because training to replicate human behavior can lead to higher interpretability. There have been many studies of these applications to fighter aircraft \cite{aircombatsurvey}. The two major classes of imitation learning are behavioral cloning and inverse reinforcement learning \cite{ILforaircombat}. In behavioral cloning, the agent attempts to mimic the exact behavior from an "expert" example, while inverse reinforcement learning seeks to find the reward function based on behavior. One example of behavioral cloning in the air domain by MIT CSAIL is the use of a novel class of algorithm called a Liquid Time Constant (LTC) to navigate a field using imagery data \cite{robustnavLTC}. \par
Reinforcement learning presents exciting advantages as opposed to classical and behavioral cloning techniques. Instead of explicit instruction to the agent, a reward function awards the agent points and penalties based on the actions it takes within an environment \cite{javorsek}. Typically this teaches the agent to operate in a more diverse range of situations. One way to leverage RL in air combat is by using hierarchical reinforcement learning, where a complex air-to-air engagement is broken down into a number of simpler tasks. Each primitive task task has its own reward function and a "policy selector" determines which task to use at each time step. 

\subsection{Sensing in Contested Environments}
 Manned fighter aircraft are equipped with a suite of sensors to give the pilot situational awareness. In the case of VENOM, the aircraft may include an Electronic Warfare (EW) suite, Radar Warning Receiver (RWR), Active Electronically Scanned Array (AESA), and an Infrared Search and Track (IRST) \cite{vipershield}. Suites can vary widely between different models, and often within different iterations (blocks) of the same model. \par
In live air combat, the environment and the adversary will challenge the ability of these sensors to accomplish the mission. RWRs can present false positive threat indications. AESA radar are subject to range and velocity ambiguity associated based on the PRF of the radar \cite{prf}. IRST cameras may suffer from dropout as a result of lack of contrast in the sky. In contested environments, loss of GPS data may cause the aircraft to rely upon its INS which can often have drift rates of 0.8 nm/hr or worse \cite{insdrift}. As US warfare transitions from permissive to contested environments, one of the most impactful consequences to operational capability is the lack of continuous GPS positioning data \cite{gpsjambad4strategy}. Factors such as ranges, field of regard, slew rate, wavelength, power can vary between sensors and as a result of software or hardware changes. In short, a rigid approach to accurately replicating detection performance in a rapidly evolving software dependent project is a losing strategy. 

\subsection{Simulations vs Training Environments}
A training environment presents fundamentally different capabilities than those of a simulation. Simulations excel in producing environments that allow humans to see similarities with the real world. However, the effectiveness of simulations as a research tool can be limited by constraints imposed to produce these similarities at low cost. An example is Harfang3D, which offers state of the art open source air combat simulation but is limited to air-to-air engagements \cite{harfang}. Furthermore, aspects that can help humans draw connections to reality may be irrelevant for helping machines. This could include: cockpit layout, motion sensing, and depth perception. By contrast, training environments are suited to fundamental research because they can be more easily customized. Meta's Nocturne project uses the Waymo dataset to present self-driving agents with a variety of scenes, traffic behaviors and sensing capabilities \cite{nocturne2022}. This level of variable isolation is not available in any known high performance aircraft simulator. \par      
The need for using training environments as well as simulators will increase as autonomous air combat continues to evolve. DARPA ACE's progress in autonomy has been groundbreaking. However, it was only possible through tens of millions of dollars of investment and years of specialized software and hardware operated by hundreds of subject matter experts. Congress has challenged both the Replicator, an initiative to deliver large numbers of small Unmanned Aerial Systems (sUAS), and CCA programs to be delivered within a matter of years \cite{replicator}. This means that autonomous air combat will be managed by testers and operators, to whom research is not a main priority. This does not mean fundamental exploration of novel ML techniques can be ignored. In fact, as described by the Chief of Staff of the Air Force, the Air Force must change to respond to rapid development of technology \cite{csafcase4change}. Tunnel presents a means to strengthen this bridge by giving a wider range of ML researchers access to high performance flight dynamics and offering military decision makers an understandable primitive to respond to evolving operational needs.   

\section{Design}
\subsection{Features}

\begin{wrapfigure}[10]{r}{0.4\textwidth}
    \centering
    \includegraphics[width=0.4\textwidth]{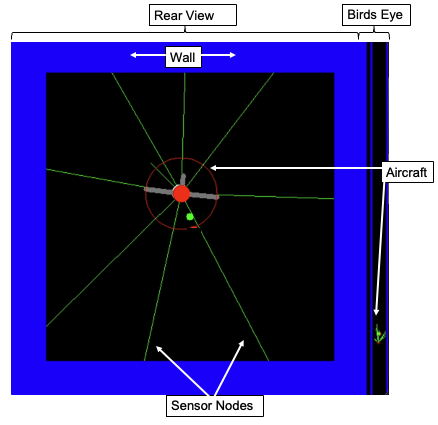}
    \caption{Tunnel Training Environment}
    \label{fig:simplelabelled}
\end{wrapfigure}

Tunnel uses Gymnasium's standardized format for constructing the initialization, reset, and step of the environment at each time step. Figure \ref{fig:simplelabelled} shows the environment. The design features are listed below:\par 

\begin{enumerate}
\item Non-linear 3D F-16 dynamics.
\item Dynamic and consequential task.
\item Open source.
\item Simple Python code.
\item Customizable state, action and rewards.
\item Customizable sensors. 
\item Primitive for non air-air missions.
\item Standardized with ML community.
\end{enumerate}
\par

\subsection{Description}\label{env_desc}

\textbf{Orientation} - The agents goal in the Tunnel is to reach the end without hitting the wall. The "end" is defined as a north position approximately 1.5 nautical miles from the start. Within the rear view, left/right represents west/east movement and up/down represents increases and decreases in altitude. The birds eye view on the far right portion of the screen shows the agents progress along the Tunnel by representing an increase in northerly progress with an upward movement on the screen.\\
\textbf{Aircraft} - The agent controls the aircraft. The exhaust of the aircraft is represented by a filled red circle. Grey lines extend spanwise to represent the wings, as well as a vertical tail. The aircraft will roll and pitch in accordance with 3D non-linear flight dynamics \cite{yechout}. \\
\textbf{Sensor} - In the default configuration, the sensors are arranged from -45\textdegree to 45\textdegree in body axis elevation and azimuth each 3\textdegree. The sensors return the distance from the center of the aircraft to the closest wall in the direction of each node. This could be thought of as a return from a LiDAR. The environment can be customized to represent a wide range of sensors, which is demonstrated in section VI. \\
\textbf{Boundaries} - As the agent progresses farther north, there is a constant width and height that is four times the wingspan of the aircraft. The full tunnel length replicates about 1.5 nautical miles of distance. The unfilled red circle around the aircraft is the aircraft radius as encoded in the environment. Any trespass of the red circle to the blue borders (walls) will cause a reset of the environment.\\
\textbf{Observations} In addition to returns from the sensor data, the environment provides access to a 16 element vector of the aircraft's state. This includes roll, yaw, pitch, airspeed, position and more. \\
\textbf{Actions} - The agent is able to control the F-16 through controls that would typically be available to the pilot. This includes control stick up/down/left/right as well as throttle and rudder actuation. \\

\section{Trade Study}

Following is a demonstration of the speed that users can expose the agent to new observations, actions, training methodologies and tasks. It is not meant to be complete or prescriptive for future developers. Instead, the hope is that it inspires collaboration and demonstrates the theoretical ability of Tunnel to rapidly adjust to changes in sensors and missions. 

\subsection{Reinforcement Learning}

Because it leverages the Gymnasium standardized architecture, Tunnel is well suited to pursuing reinforcement learning training. The following study investigated the extent to which the agent could control the aircraft. In this study, "effectiveness" and "success" are defined by the agents' ability to reach the end of the Tunnel.  \\
\textbf{Hypothesis} - The hypothesis was that an RNN could navigate the Tunnel more effectively than an MLP when trained via Proximal Policy Optimization (PPO).\\
\textbf{States} - This demonstrated one instance of rapid iteration of different observation spaces. The initial space included the array of distances to the nearest wall provided by the sensor as well as the aircraft's internal state, NX, as described in Section \ref{env_desc}. Different aspects of NX were removed from the observation space as experimentation continued. Different iterations investigated different azimuth and elevation bounds as well as changed the number of sensor nodes. A 3 x 3 vector at -60\textdegree, 0\textdegree and 60\textdegree in elevation and azimuth relative to body axis was the configuration used for training. Various histories of the sensor data was provided as well, with the final iteration giving the agent an observation space of the last four timesteps of the sensor returns and no internal state data.    \\
\textbf{Actions} - Some exploration of isolating action space was conducted to troubleshoot navigation throughout the training process. The action space was the default continuous action space described in Section \ref{env_desc}. \\
\textbf{Rewards} - The first iteration gave the agent a penalty for the distance from the center of the Tunnel. Based on the results of the "Trackmania" effort~\cite{trackmania}, the reward function was altered to a construct that rewarded the agent for passing "targets" within the Tunnel. The targets were aligned within the center of the Tunnnel at regular intervals of North position. Any time the agent's North position passed the target it would receive the reward regardless of deviation from the center. The final iteration involved an arithmetic series of target rewards, IE: 100, 200, 300, ..., 38000.   \\
\textbf{Findings} - The agent was less successful in navigating the Tunnel for both RNN and MLP models when the aircraft's state and the sensor data were both part of the observation space. Using sensor data alone yielded better results. Though neither reliably navigated the Tunnel. There was no significant difference seen between RNN and MLP agents. 

\subsection{Custom Observations \& Actions}

\begin{wrapfigure}[17]{r}{0.7\textwidth}
    \centering
    \includegraphics[width=0.7\textwidth]{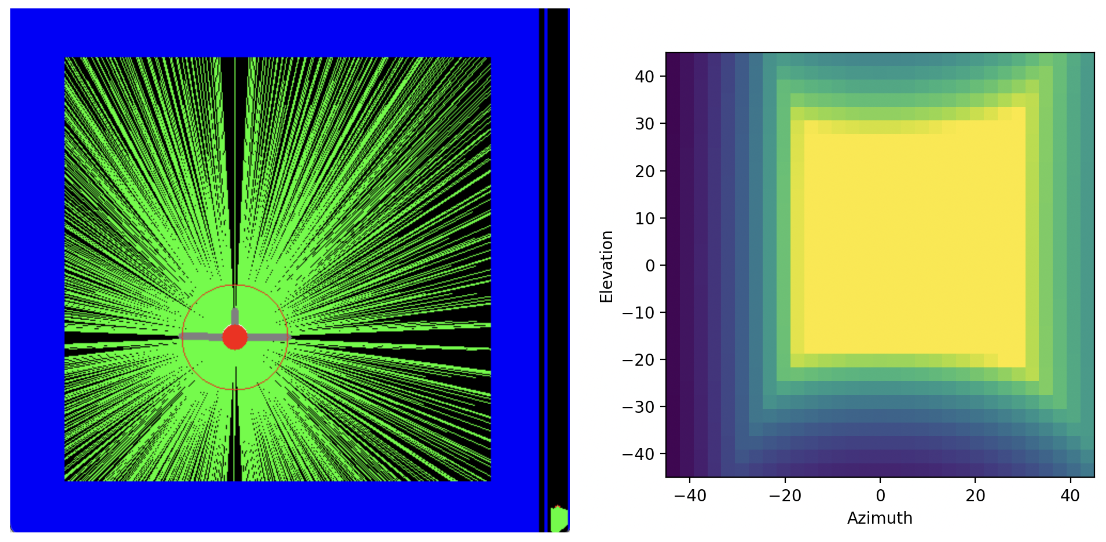}
    \caption{Imagery Input: LiDAR returns at each node colored by distance}
    \label{fig:echomap}
\end{wrapfigure}

The simple and open source nature of Tunnel allows developers to make rapid low-level changes to the environment to adapt to evolving sensor and mission needs. \\
\textbf{Setup} - The density of sensor nodes was increased to one per 3\textdegree and decreases the field of regard to -45\textdegree to 45\textdegree for azimuth and elevation. As can be seen in Figure \ref{fig:echomap}, this gives the aircraft an "image" of ranging data. In this case, because the aircraft is on the lower and western portion of the Tunnel, there is a higher range return for azimuth right of the aircraft center-line and elevation above the aircraft's nose. To increase sample diversity, the aircraft was initialized with a random angle between 0\textdegree and 360\textdegree during training. The agent's initial position would be displaced 150\% the aircraft's' radius at this random angle. \par
Instead of using reinforcement learning, imitation learning was used to train the agent.. This showed that Tunnel can be used for training schemes beyond reinforcement learning. Training data was obtained from an expert model. Typically this would be a human. For data clarity, a waypoint-following autopilot was used to control the aircraft to the center of the Tunnel. This autopilot, provided by a team at Air Force Research Lab, has the F-16 3D non-linear dynamics in the back end. The autopilot takes aircraft state and waypoint as input and outputs a control in the form of the default Tunnel action space. The waypoint was placed at the max depth of the tunnel in the center of the height and width. Also, the training occurred only with only control stick inputs. Meaning, throttle and rudder were not factored into the training. With Tunnel, this adjustment was trivial. Instead of the LiDAR image, each network was given an array of zeroes of similar shape. 

\textbf{Findings} - A very simple PID "expert" was able to navigate the Tunnel reliably. Equations for this controller can be found in the Appendix. The autopilot "expert" could navigate the Tunnel as well with far less deviation from the centerline of the Tunnel.   \par 
Although the PID controller was able to reliably navigate to the end of the tunnel, training an agent to do the same via behavioral cloning was unsuccessful. The agent trained on the PID expert via behavioral cloning underperformed the PID expert. The PID controller as well as the agent trained through PID were far more successful with longitudinal (pitch) control than lateral-directional (yaw/roll) control. This may be related to the F-16s higher stability in the longitudinal axis than the lateral-directional. It shows that in this case an important factor to training success was stability. A lack of stability increased time sensitivity of the agents actions and caused a lower range of acceptable parameters for tuning. Errors in parameter tuning compounded as the agent was trained via behavioral cloning.

\subsection{Mission and Sensor Modifications}

\begin{wrapfigure}[25]{r}{0.5\textwidth}
    \centering
    \includegraphics[width=0.5\textwidth]{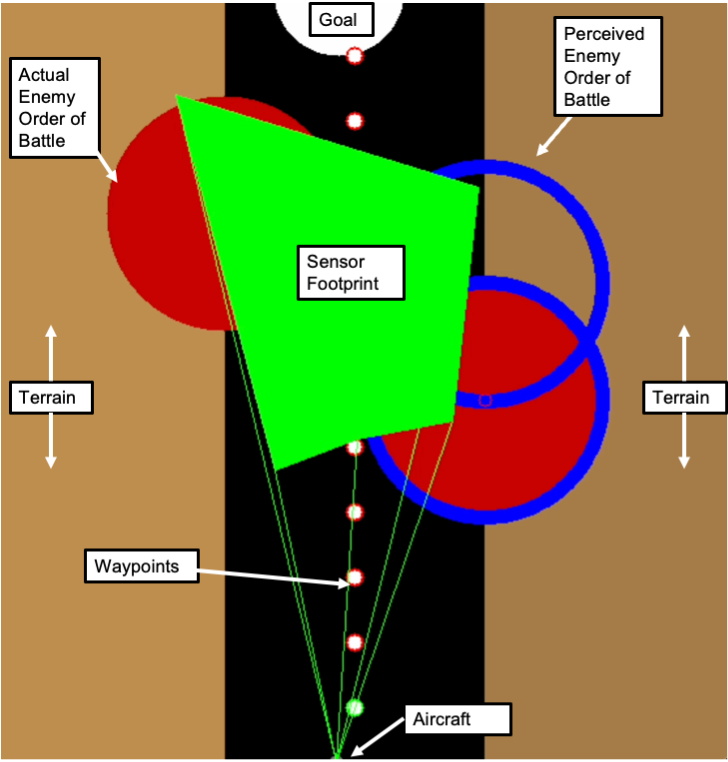}
    \caption{Missionized Tunnel Environment}
    \label{fig:desert}
\end{wrapfigure}

Next is a simple modification to Tunnel that begins to represent an operational mission. We call this a "missionized" Tunnel environment. The following alterations were possible within a week with a team of one. \\
\textbf{Orientation} - Figure \ref{fig:desert} depicts the new training environment. This is a top down view of the aircraft navigating through a more complex Tunnel. Adversary missile engagement zones are shown as red circles. The blue unfilled circles are the agent's perception of the enemy as recorded by an out of data enemy order of battle. The green polygon is the ground track of the forward looking sensor. Brown boundaries are terrain of varying heights. The white circle at the north end of the environment is the goal, meant to represent a safe point.\\
\textbf{Mission} - The agent must navigate to the goal region while avoiding adversary engagement zones. This is meant to represent a primitive task that a autonomous high performance aircraft may be asked to perform within a partially observed environment. \\
\textbf{Targets} - The navigation path was updated to require the agent to generate targets onboard using the perceived enemy order of battle. The agent plots the most efficient course to the goal while avoiding threats using the A* algorithm \cite{astars}.   \\

\textbf{Sensors} - This scenario forces the agent to navigate without use of GPS data The aircraft collects distance returns from an omnidirectional, short range laser range finder. In addition, as depicted by the green polygon, it has a long range forward looking camera. This could represent a FLIR or optical sensor. Note that although using reinforcement learning to navigate to waypoints was difficult, this task was reliably accomplished by use of an autopilot. \\

\textbf{Adversary} - The Gymnasium training environment template allows for adversarial training. In this case, the adversary's action space is a on/off for both missile engagement zone. With the threat turned on, the adversary can sense the agent, but is visible to the agent. This could motivate a self-play reinforcement learning situation. \\

\textbf{Findings} - There were instances where the agents' sensor did not discover an updated enemy order of battle which caused trespasses of the missile engagement zone. If sensor discovery was no factor, the the agent was able to reliably navigate via use of an autopilot. This extension on the baseline Tunnel was completed in less than a week.

\section{Discussion}

\subsection{Training Approach Comparison}
This study has shown an example of how different training approaches can be assessed quickly using training environments that leverage dynamics of high performance aircraft. One finding from this study is that a more sophisticated machine learning algorithm doesn't necessarily increase performance. With the same observation space of LiDAR returns, behavioral cloning was far more successful in navigating the Tunnel than reinforcement learning. A principle supported by this finding is to choose the autonomy methodology that is most simple and reliable that will accomplish the task when designing autonomous systems in the air domain. \par
However, it is important to consider the importance of the rate in which agents need to respond to changes in the observation space. This can cause unpredictability which could be exacerbated by partial observability. For example, the classical PID controller was able to maintain aircraft control within a far wider range of parameter values in the longitudinal direction than in the lateral-directional regime. The lower stability in roll caused the agent to need to respond at a much higher frequency which decreased the likelihood of success of the PID controller. There was a much narrower range of acceptable parameters in the roll direction. 
\par
It's important to consider the vast difference in risk tolerance in air combat as compared to other domains for ML. While 80\% or 90\% success rate may be acceptable in consumer image recognition, failure in aviation is far less lenient. According to the US Air Force Test Center guidance, high, medium and negligible likelihood of risk is defined as one in every 10, 10 thousand, and 10 million, respectively. For this reason, the "corner cases" cannot be ignored. A potential heuristic is to choose the training approach that will lead to "corner cases" that can be most effectively mitigated based on the resources available. Table \ref{demo-table} shows a summary of findings in this limited task. Success is defined as the agent reliably navigating to the end of the Tunnel. Marginal indicates that the agent reached the end in some samples, but not reliably.   

\begin{table}[hbt!]
\caption{\label{tab:demo-table} Comparison of Training Methods}
\centering
\begin{tabular}{lcccc}
\hline
Environment & Training & Characteristic  & Success \\\hline
Template & Reinforcement Learning & Sparse sensor array & Marginal \\
Template & Reinforcement Learning & Dense sensor array & No \\
Template & Behavioral Cloning & Autopilot expert & Yes \\
Template & Behavioral Cloning & Hand build expert & No \\
Missionized & Classical & Autopilot w/ perfect EOB & Yes \\
Missionized & Classical & Autopilot w/ incorrect EOB & Marginal \\
Missionized & Classical & Autopilot w/ rapid altitude and heading changes & Marginal \\
\hline
\label{demo-table}
\end{tabular}
\end{table}

\subsection{Applicability to Real Sensors and Missions}
There were many limitations in this study's ability to replicate a particular technical solution to a tactical problem. It did not have operationally representative ranges for any particular sensor specifications. It also did not align with one mission. Instead, it showed a number of primitive capabilities that could be extended when applied to advanced research. \par
Consider the transfer to a more specialized simulation from the perspective of the agent. In the flight test regime, the walls could represent altitude and heading tolerances that are commonly used to assess navigation capability. A wall could also represent more abstract boundaries such as an airspace limitation or known missile engagement zone. Once integrated, these differences should be transparent to the agent. Though sensors were limited to LiDAR return in this study, a "picture" of the battlespace as perceived by the agent can be understood as a combination of various sensors into an array. Unlike a human operator whose comprehension of a pixel on a display relies on subject matter expertise, the agent may be able to operate using this array of sensor input alone. \par
While this study does not provide a definitive answer as to how successful autonomous aircraft will execute future air combat missions, it begins to show the connection between simple "drills" in a training environment and the real world. As autonomous air combat becomes more sophisticated, this can help break down more complex operations into tasks that can be both trainable by the agent and interpretable by the human. 

\section{Future Work}
\subsection{Deployment of Tunnel}
This paper presented a small sample of the potential that the Tunnel environment can provide when presented to a larger audience. Machine learning researchers now have an open source means to apply their vast expertise to a dynamic task with critical consequences. Air domain experts have a tool that can be adapted to a host of missions and sensor configurations. The author will continue to create connections between these domain experts and researchers. An ideal use of Tunnel may be a response to a novel training methodology or operational requirement that cannot be accurately represented in current simulation. In this case Tunnel can first be configured to the mission requirement and sensors available. Then hundreds of combinations of observation spaces, action spaces, ML algorithms, and training techniques can be evaluated against each other. Once decision makers have chosen candidate approaches, these methodologies could be integrated into a higher performance solution like JSBSim, gigastep \cite{gigastep} or proprietary software. 

\subsection{Collect High Quality Training Data}
As stated, high quality, labelled training data from fighter training missions can help make imitation learning a more viable option. Reinforcement learning has been demonstrated as a reliable and powerful tool for Air-to-Air engagements through the last years of DARPA ACE. However, as autonomous air combat begins to integrate real world sensors and a diverse mission set, efforts to simplify problem sets will be crucial. The author will work to establish a pipeline for unclassified data recorded from manned F-16 flights to be more available. 

\subsection{Future Combat Autonomy}
This study seeks to provide a tool to explore the future of autonomous air combat. Senior leaders have challenged members of the AIA to consider autonomy in the air domain in terms of a wide variety of missions in addition to fighter operations. This vision has been designated as Autonomous Combat Platforms (ACPs). Airmen have begun to answer this call with concepts of a progression of autonomy from a wingman, to a flock, to a swarm \cite{blairflocking}. These efforts help find technical solutions to DARPA's Mosaic Warfare concept, which uses flexible force compositions to enable US decision dominance while imposing the maximum ambiguity on the adversary \cite{mosaicwarfare}.

\section{Conclusion}

This has shown one example of the simplicity and speed at which trade studies can be conducted within the Tunnel environment. A missionized extension, while not high fidelity, can expose the agent to relevant observation space and high performance flight dynamics. These can be iterated within less than a week by a user without niche expertise. \par
Tools that enable rapid exploration are essential to the evolution of autonomous air combat from VISTA to CCA. Tunnel gives researchers and operators a means to collaborate and quickly modify the training environment to their particular mission needs. This paper has shown that a trade study can assess a range of observations, actions, tasks and training methodologies within a short timeline. This paper is an invitation to those looking to bridge research with operational relevance in high performance aircraft. It is through this collaboration that the potential of this effort can be realized.

\newpage

\section*{Appendix}

\subsection{PID Controller Equation}

The controller was constructed via:\\ \\
\[N_z(t) = -0.002*E_y(t-1)-0.2*dE_y(t-1)  \] \\
\[P_s(t) = -0.001*E_x(t-1)-0.1*dE_x(t-1)  \] \\
Where:\\
\[E_x = x - center_x\] \\
\[E_y = y - center_y\] \\
\[dE_x(t) = x(t) - x(t-1)\] \\
\[dE_y(t) = y(t) - y(t-1)\] \\

\section*{Acknowledgments}
This work was sponsored by the DAF AI Accelerator. I am grateful for the opportunity to pursue research that aligns with both my passions and work within the Air Force. I am humbled by the members of my Phantom cohort, C-10 aka C-X, who are a constant source of inspiration. Thanks also to Maj Joshua Rountree, Dr. Ross Allen and the rest of the Lincoln Labs team: Jaime Pena, Rodney Lafuente Mercado, Kaise Al-Natour and William Li for contributing to the Air Guardian Autonomy project.

\bibliography{sample}

\end{document}